\documentclass[10pt, conference]{IEEEtran}
\IEEEoverridecommandlockouts
\usepackage[utf8]{inputenc} 
\usepackage{url} 
\usepackage{booktabs} 
\usepackage{nicefrac} 
\usepackage{microtype} 
\usepackage{cite}
\usepackage{amsmath,amssymb,amsfonts}
\DeclareMathOperator*{\argmax}{arg\,max}

\usepackage{algorithmicx}
\usepackage{algorithm}
\usepackage[noend]{algpseudocode}
\usepackage{graphicx}
\usepackage{textcomp}
\usepackage{xcolor}
\usepackage{subcaption}
\usepackage[super]{nth}
\usepackage[numbers]{natbib}
\usepackage{etoolbox}
\def\BibTeX{{\rm B\kern-.05em{\sc i\kern-.025em b}\kern-.08em
T\kern-.1667em\lower.7ex\hbox{E}\kern-.125emX}}
\begin{document}

\title{Efficient Multivariate Bandit Algorithm \\ with Path Planning}

\makeatletter
\newcommand{\newlineauthors}{%
\end{@IEEEauthorhalign}\hfill\mbox{}\par
\mbox{}\hfill\begin{@IEEEauthorhalign}
}
\makeatother
\newcommand{\ts}{\textsuperscript}
\newcommand{\shortauthors}{Keyu and Zezhong, et al.}
\newcommand\scalemath[2]{\scalebox{#1}{\mbox{\ensuremath{\displaystyle #2}}}}

\author{
\IEEEauthorblockN{Keyu Nie \textsuperscript{\textsection}}
\IEEEauthorblockA{\textit{eBay Inc.}\\
San Jose, USA\\
keyunie@gmail.com}
\and
\IEEEauthorblockN{Zezhong Zhang \textsuperscript{\textsection}}
\IEEEauthorblockA{\textit{eBay Inc.}\\
San Jose, USA\\
zezzhang@ebay.com}
\and
\IEEEauthorblockN{Ted Tao Yuan}
\IEEEauthorblockA{\textit{eBay Inc.}\\
San Jose, USA\\
teyuan@ebay.com}
\and
\IEEEauthorblockN{Rong Song}
\IEEEauthorblockA{\textit{eBay Inc.}\\
San Jose, USA\\
rsong@ebay.com}
\and
\IEEEauthorblockN{Pauline Berry Burke}
\IEEEauthorblockA{\textit{eBay Inc.}\\
San Jose, USA\\
pmburke10@gmail.com}
}

\maketitle

\begingroup
\renewcommand\thefootnote{\textsection}
\footnotetext{Both authors contributed equally to this research.}
\endgroup

\begin{abstract}
In this paper, we solve the arms exponential exploding issues in the multivariate Multi-Armed Bandit (Multivariate-MAB) problem when the arm dimension hierarchy is considered. We propose a framework called path planning (TS-PP), which utilizes paths in a graph (formed by trees) to model arm reward success rate with m-way dimension interaction and adopts Thompson Sampling (TS) for a heuristic search of arm selection. Naturally, it is straightforward to combat the curse of dimensionality using a serial process that operates sequentially by focusing on one dimension per each process. For our best acknowledge, we are the first to solve the Multivariate-MAB problem using trees with graph path planning strategy and deploying alike Monte-Carlo tree search ideas. Our proposed method utilizing tree models has advantages comparing with traditional models such as general linear regression. Real data and simulation studies validate our claim by achieving faster convergence speed, better efficient optimal arm allocation, and lower cumulative regret.
\end{abstract}


\section{Introduction}
The Multi-Armed Bandit (MAB) problem \cite{thompson1933, Lai&Robbins1985}, is widely studied in probability theory and reinforcement learning. Binomial bandit is the most common bandit formats with binary rewards ($X^{t} \in \{0, 1\}$). The upper confidence bound (UCB) algorithm was demonstrated as an optimal solution to manage regret bound in the order of $O(log(T))$\cite{Lai&Robbins1985},
 where $T$ is the number of repeated steps. In an online system, Thompson Sampling (TS) algorithm \cite{thompson1933} attracts lots of attention \cite{google2012}
 due to its simplicity and efficiency but resistance in batch updating when comparing with UCB. 

Modern applications (e.g., UI Layout) have configuration involving multivariate dimensions to be optimized, such as font size, background color, title text, module location, item image, and each dimension contains multiple choices \cite{amazon2017, nair2018finding}. The exploring space (arms) faces an exponential explosion of possible option/value combinations in dimensions. In this paper, we call it the Multivariate-MAB problem, and it has applications in other areas as well. Thompson Sampling solutions \cite{google2012}
 are slowly converged to optimal arm \cite{amazon2017} under the curse of dimensionality. \cite{amazon2017} proposed the \textbf{MVT} utilizing hill-climbing optimization \cite{casella2002statistical} within dimensions, and reported that it significantly reduced the parameter space from exponential to polynomial. However, in our replication, we found that updating regression weights at each iteration created a heavy computation burden.

\begin{figure}[htbp]
\centering
\begin{subfigure}[b]{\columnwidth}
\includegraphics[width=\columnwidth, scale = 0.1]{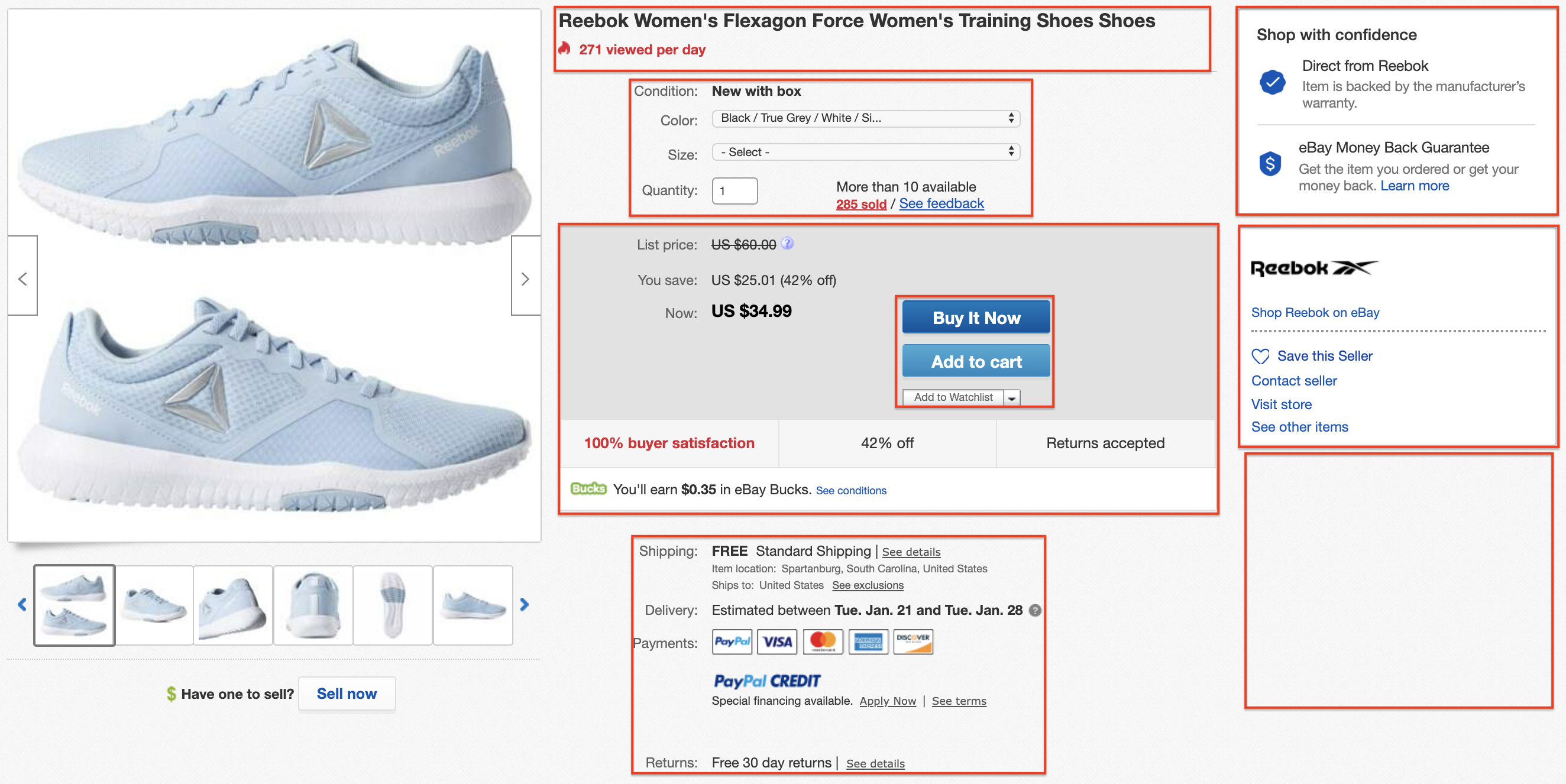}
\end{subfigure}
\caption{Example of eBay view-item page to demo Multivariate-MAB problem. Each rectangle box is a dimension containing several choices to be optimized.}
\label{MultivariateMAB}
\end{figure}

We propose a framework "Path Planning" (TS-PP) to overcome these shortcomings. Our novelty includes: \textbf{(a)} modeling arm selection procedure under tree structure; \textbf{(b)} efficient arm candidates search strategies; \textbf{(c)} remarkable performance improvement by straightforward but effective arm space pruning. This paper is organized as follows: firstly introduce the problem setting and notation; then explain our approach in detail, and further discuss the differences among several variations; also examine the algorithm performance in simulated study and conclude at the end.

\section{Multivariate-MAB Problem} \label{sec2}
We start with the formulation of multivariate-MAB: the decision of layout (e.g., web page) $A$, which contains a template with $D$ dimensions, and each dimension contains $N$ choices, to minimize the expected cumulative regret $R$. The binary reward $X \in \{0, 1\}$ is discussed in our paper, but we would extend to categorical/numeric rewards in our future work. We assume the same number of choices for all dimensions for simplicity purposes, and it should not affect our claims. The layout $A$ is represented by a D-dimensional vector $[f_{1}, \dots, f_{D}] \in \{1,\dots,N\}^{D}$. $A[i] = f_i$ denotes the selected content for the i-th dimension.

At each step $t = 1, \dots, T$, bandit algorithms selects arm $A^{t}$ from $N^D$ choices over search space to minimize cumulative regret over $T$ rounds:
\begin{align*}
\scalemath{0.9}{
R_{T} =\sum_{t = 1}^{T} ( E(X | A^{\star t}) - E(X | A^{t} ) )
}
\end{align*}
where $A^{\star t}$ stands for best possible arm at step $t$. Generally, $R_{T}$ is on the order $O(\sqrt{T})$ under linear payoff settings \cite{dani2008stochastic, chu2011contextual, agrawal2013thompson}. We would report contextual multivariate MAB in future work. 

\section{Related Work}
A straightforward solution to solve the multivariate-MAB problem is to apply Thompson Sampling (TS) \cite{Kaufmann2012} directly over $N^D$ arms/layouts, which treat the multivariate MAB problem as the traditional MAB problem \cite{google2012, google2014} with $N^D$ arms/layouts. However, the $N^D$-MAB suffers the curse of dimensionality, and the performance worsened as the number of layouts grew exponentially \cite{amazon2017}. To overcome the exponential explosion of parameters, D-MABs \cite{amazon2017} considers the dimensions independent assumption and independently applies Thompson Sampling (TS) over $N$ arms for each $D$ dimensions, which ignores all the interactions among dimensions. To consider higher-order dimension interactions, MVT algorithms \cite{amazon2017} leveraged the probit regression model to learn the interactions among dimensions and present good performance on amazon's data with MVT2 (a special MVT considering $2^{nd}$ order interactions). 

Besides ideas for modeling layout rewards, there are other approaches to construct the layout. Though hill climbing is a heuristics method, it works well on constructing a better layout from an initial layout \cite{amazon2017}. Inspired by another heuristic method Monte Carlo tree search (MCTS) \cite{browneMCTSsurvey}, it is intuitive to construct a layout with a sequence of dimension optimization decisions. Compared with D-MABs, the Naive MCTS idea considers higher-order interactions in each optimization decision. 
\section{Proposal Solution}\label{sec3}
\subsection{Path Planning Framework}
In this paper, we propose TS-PP for the Multivariate-MAB problem. The key idea is to optimize dimensions in sequential order, with one dimension each time, while keeping other dimension content fixed. 

Inspirited by MCTS idea, TS-PP includes two steps: selection and back-propagation. At selection step, it picks a sequence of dimensions with random order $[d_1:d_{D}]$, which is any permutation of D dimensions ($[1, 2, ..., D]$). Here symbol `$:$' in $[d_1:d_{D}]$ is compact way to write $[d_1, \dots, d_{D}]$. Within the sequence $[d_1:d_{D}]$, we utilize optimizing mechanism to obtain optimized content $f^{*}_{d_i}$ for dimension $d_i$ while fixing content choices $f^{*}_{d_1:d_{i-1}}$ in pre-visited dimensions $[d_1:d_{i-1}]$ as $f^{*}_{d_i} | f^{*}_{d_1:d_{i-1}} = \argmax_{f} f_{d_i} | f^{*}_{d_1:d_{i-1}} $. The superscript $*$ indicates the content is optimized. Following this process, we construct the optimized layout $A^{c} = [f^{*}_{d_1:d_{D}}]$ via $f^{*}_{d_1} \rightarrow f^{*}_{d_2} | f^{*}_{d_1} \rightarrow f^{*}_{d_3} | f^{*}_{d_1}, f^{*}_{d_2} \rightarrow ... \rightarrow f^{*}_{d_i} | f^{*}_{d_1:d_{i-1}} \rightarrow ... \rightarrow f^{*}_{d_D} | f^{*}_{d_1:d_{D-1}}$, where $[f^{*}_{d_1:d_{D}}]$ stands for vector $[f^{*}_{d_1}, f^{*}_{d_2}, ..., f^{*}_{d_D}]$. After the reward collection, the back-propagation step updates the states of all predecessors. The description above reflects how naive MCTS performs in our framework, and we call it Full Path Finding (FPF) in this paper.

In fig.~\ref{dg}, we draw $D \times D$ nodes to present TS-PP in the graph view. There are $D$ levels in the graph, and each level contains $D$ nodes. Here a node represents a dimension $d_i$. Please keep in mind that, within each note, it contains $N$ choices denoting as $f^{1}_{d_i},...,f^{N}_{d_i}$. A sequential dimension with random order represents a path from the top (root) to bottom with non-repeated dimensions. Notably, there are $D!$ unique paths in the graph. The optimization along a `path' follows stochastic manner, which is saying that the mechanism of optimizing ($f^{*}_{d_i}$) current node $d_i$, out of $N$ choices, is highly dependent on optimized nodes lying in pre-path of the current node ($f^{*}_{d_1},...,f^{*}_{d_{i-1}}$).

\begin{figure}[htbp] 
\centering
\begin{subfigure}[b]{0.7 \columnwidth}
\includegraphics[width=\columnwidth, scale = 0.1]{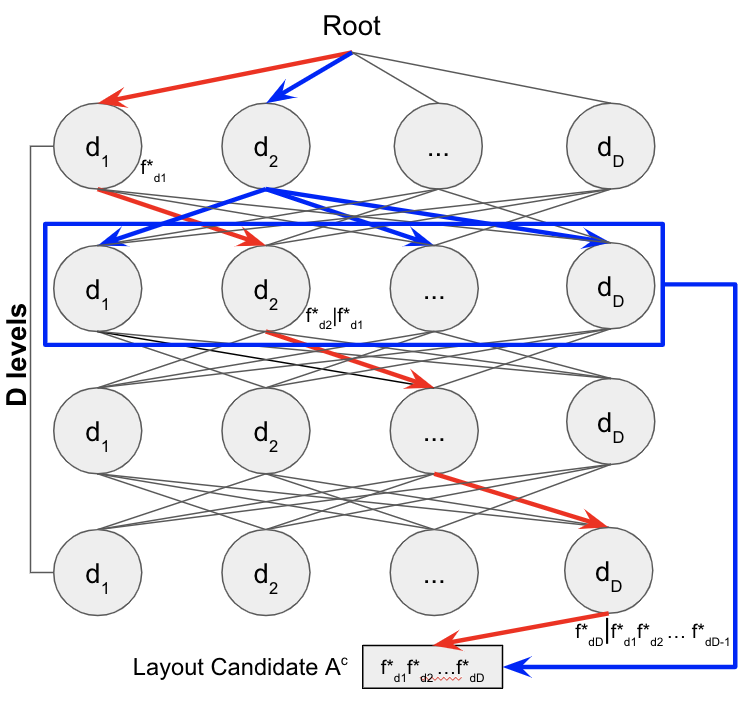}
\caption{FPF and PPF procedure}
\label{dg1}
\end{subfigure}
\begin{subfigure}[b]{0.7 \columnwidth}
\includegraphics[width=\columnwidth, scale = 0.1]{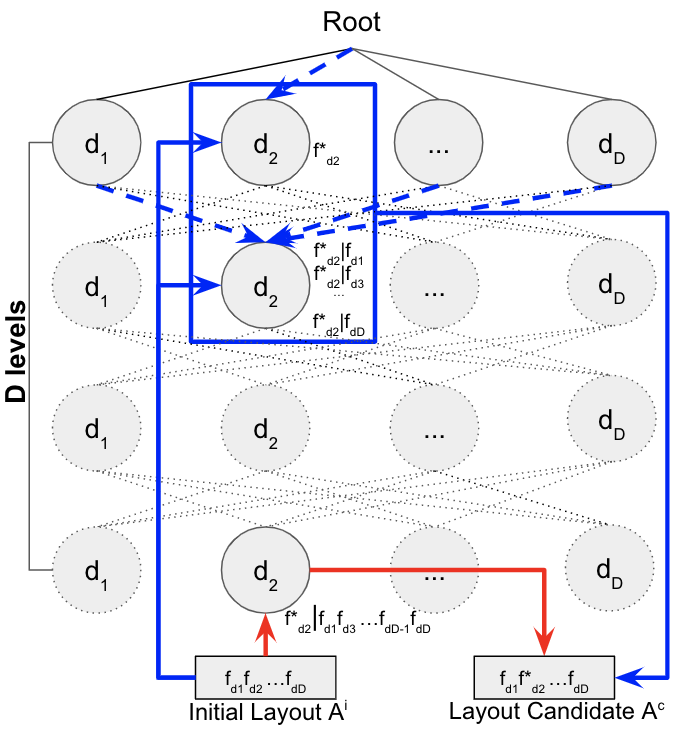}
\caption{DS and Boosted procedure}
\label{dg2}
\end{subfigure}
\caption{Graph View of Path Planning Procedure}
\label{dg}
\end{figure}

\subsection{Thompson Sampling}
Thompson sampling (TS) \cite{russo2018tutorial} utilizes Bayesian posterior distribution of reward, and selects arm proportional to probability of being best arm. Usually the binary response assumes $\mathtt{Beta}(\alpha_0, \beta_0)$ for arm $k$ as prior to compose the posterior distribution $\mathtt{Beta}(\alpha_k+\alpha_0, \beta_k + \beta_0)$, where $\alpha_k$ and $\beta_k$ are the number of successes and failures it has been encountered so far at arm $k$. The $(\alpha_k+\alpha_0, \beta_k + \beta_0)$ are hidden states associated with arm $k$. At selection stage, in round $t$, it implicitly selects arm as follows: simulate a single draw of $\theta_k$ from posterior ($\mathtt{Beta}(\alpha_k+\alpha_0, \beta_k + \beta_0)$) for each arm $k$ and the arm $k^{*} = \argmax_k(\theta_k)$ out of all arms will be selected. 

\begin{algorithm}[htbp]\footnotesize
\caption{Implicit Sampling} \label{ts}
\begin{algorithmic}[1] 
\Function{Sample}{Fixed=$[f_{d_{1}:d_{i}}]$} 
\State {$
\begin{aligned}
\text{Get} \; \alpha, \beta \; &\text{from} \; \textbf{Node}(f_{d_{1}:d_{i}})
\end{aligned}
$}
\State sample $\theta \sim \mathbf{Beta}(\alpha + \alpha_0, \beta + \beta_0)$
\State \textbf{Return} $\theta$
\EndFunction 
\end{algorithmic}
\end{algorithm}

Algorithm \ref{ts} shows pseudo code on how to implicitly sample $\theta_k$ described in TS, which is heavily adopted in TS-PP too. The function $\textbf{Node}(f_{d_{1}:d_{i}})$ returns the hidden states of $f_{d_{1}:d_{i}}$, and it assembles two components together: content $f_{d_{i}}$ for target/current node $d_{i}$ at level $i$ as well as predecessor nodes ${d_{1}:d_{i-1}}$ with fixed contents $f_{d_{1}:d_{i-1}}$. At back-propagate step, \textbf{Node} would be updated with the collected reward $X^{t}$. \footnote{In practice, $\textbf{Node}(f_{d_{i}} | f_{d_{1}:d_{i-1}})$ requires $O(1)$ computation complexity. But it could also be implemented in $O(T)$ computation complexity for lazy back-propagation with cache memory saving. }
The hidden states is associated with $f_{d_{1}:d_{i-1}}$, not $f_{d_{i}} | f_{d_{1}:d_{i-1}}$. $f_{d_{i}}$ and $f_{d_{1}:d_{i-1}}$, the split of $f_{d_{1}:d_{i-1}}$, are interchangeable and they represents the same joint distribution (with same hidden states) of $f_{d_1:d_{i}}$. We still utilize a particular example in fig.~\ref{dg1} red arrows for explanation. Directed edges in fig.~\ref{dg} connect parent node to child node in the graph, where the arrow represents conditional (jointly) relationship. At level $i$ along the path in red arrow, the content $f_{d_i}$ within node $d_i$ plus the predecessor nodes with choices $f_{d_1:d_{i-1}}$ (the prefixed of $f_{d_i}$) is a conditional probability $P(f_{d_i} | f_{d_1:d_{i-1}}) \propto P(f_{d_1:d_{i}})$. In practice, $P(f_{d_i}, \text{preFixed}(f_{d_i})) =P(f_{d_1:d_{i}})$ could be represented by hidden states $(\alpha, \beta)$ from Beta distribution (binary rewards) for content $f_{d_i}$ within node $d_i$ (given its $\text{preFixed}(f_{d_i})$). The intuition that this stochastic manner would work is straight forward. Since the likelihood of a particular layout $A = [f_{d_1:d_{D}}]$ being the best arm (out of $N^D$) depends on its posterior distribution $P(f_{d_1:d_{D}})$. Based on the chain rule, $P(f_{d_1:d_{D}}) = \prod_{i = 1}^{D} P(f_{d_i} | f_{d_1:d_{i-1}}) \propto P(f_{d_i} | f_{d_1:d_{i-1}}) \propto P(f_{d_1:d_{i}})$, which means it is also partially related with $P(f_{d_1:d_{i}})$. Instead of sampling directly from posterior distributions of $N^D$ arms ($f_{d_1:d_{D}}$), sampling stochastically from distribution $f_{d_1:d_{i}}$ could also provide guidance on value selection for dimension $d_i$.

\subsection{TS-PP Template}
Algorithm~\ref{framework} provides the TS-PP template for a better understanding of our proposal over the big picture. In the selection step, The proposed algorithms utilize different path planning procedures to obtain candidate arm ($A^{s}$) at the selection stage. Path planning procedure navigates the path from predecessor node to successor node originating from start to finish in the graph and applies TS within a target node (dimension) to optimize the best content condition on fixing optimized contents in visited predecessor nodes unchanged. To avoid stack in sub-optimal arms under this heuristic search, we intentionally repeat our search ($S$) times and re-apply TS tricks among these $S$ candidates for final arm selection.: $\argmax_{A^{s}}(\theta^{s})$. At back-propagate step, the history reward $\mathbf{H}^{t}$ and hidden states in \textbf{Node}() are updated.

\begin{algorithm}[htbp]\footnotesize
\caption{TS-PP Template} \label{framework}
\begin{algorithmic}[1] 
\State \textbf{Input} D, N, S $\in$ $R^1$, $\alpha_0 = 1$, $\beta_0 = 1$
\For {\text{step} $t = 1, 2, \dots$}
\For {\text{search} $s = 1 \to S$} 
\Comment {Selection Step}
\State $A^{s} \gets$ Path Planning Procedure
\State Sample $\theta^{s} \gets \mathbf{Sample}($Fixed$ = A^{s})$ \label{sample_theta}
\EndFor
\State Select Arm $A^{t} \gets \argmax_{A^{s}}(\theta^{s})$
\State Update History $\mathbf{H}^{t} \gets \mathbf{H}^{t-1} \cup (X^{t}, A^{t})$ \Comment {Back-propagate Step}
\State Update \textbf{Node}() with $\mathbf{H}^{t}$
\EndFor
\end{algorithmic}
\end{algorithm}

\subsection{Path Planning Procedure}
We propose four path planning procedures for candidates searching: Full Path Finding (\textbf{FPF}), Partial Path Finding (\textbf{PPF}), Destination Shift (\textbf{DS}) and Boosted Destination Shift (\textbf{Boosted-DS}). \textbf{FPF} and \textbf{DS} are direct application of MCTS and hill-climbing with top-down and bottom-up strategies for layout selection respectively. Under graph view, \textbf{FPF} (fig.~\ref{dg1} red arrow) employs the depth-first search (DFS). \textbf{PPF} (fig.~\ref{dg1} blue arrow) replaces DFS strategy with breath-first search (BFS) strategy for enhancement. \textbf{Boosted-DS} (fig.~\ref{dg2} blue arrow) enhances \textbf{DS} (fig.~\ref{dg2} red arrow) by sampling from top nodes instead of bottom. The following discusses the four procedures with details.

\textbf{Full Path Finding:} 
\textbf{FPF} is the direct application of MCTS and describes a DFS algorithm in graph search. The details is specified in pseudo code in algorithm~\ref{alg2} from line~\ref{full} to \ref{fullend}. $[d_{1}:d_{D}]$ is a permutation of dimensions $[1:D]$ and is initialized completely randomly with equal chance. Starting from top ($d_{i}$), \textbf{FPF} recursively applies TS policy following order of $[d_{1}:d_{D}]$ to obtain optimized layout $A = [f^{*}_{1:D}]$. The computational complexity is $O(SND)$ with $S$ repeated searches, and space complexity is $O(N^D)$. For lazy back-propagation, the computation and space complexity could be improved to $O(SNDT)$ and $O(SDT)$ separately.

\begin{algorithm}[htbp]\footnotesize
\caption{Path Planning Procedures} \label{alg2}
\begin{algorithmic}[1] 
\Procedure{\textbf{Full Path Finding}}{} \label{full}
\State Initial $[d_{1}:d_{D}]$ randomly
\For {$d_{i} = d_{1} \to d_{D}$}
\State $f^{*}_{d_{i}} \gets \mathbf{TS}$(tgtDim=$d_{i}$, Fixed=[$f^{*}_{d_{1}:d_{i-1}}$])
\EndFor
\State Return $A = [f^{*}_{1:D}]$
\EndProcedure \label{fullend}
\Procedure{\textbf{Partial Path Finding}}{} \label{partial}
\State Choose $d_i \in [1:D]$ randomly
\State $f^{*}_{d_i} \gets \mathbf{TS}($tgtDim$=d_i$, Fixed$=\emptyset)$
\For {$(d_{j} = d_{1} \to d_{D})$ and ($d_{j} \neq d_{i}$)}
\State $f^{*}_{d_j} \gets \mathbf{TS}$(tgtDim=$d_j$, Fixed=[$f^{*}_{d_j}$])
\EndFor
\State Return $A = [f^{*}_{1:D}]$
\EndProcedure \label{partialend}
\Procedure{\textbf{Destination Shift}}{} \label{dest}
\State Initial $A$ randomly
\For {$k = 1 \to K$}
\State Choose $d_i \in [1:D]$ randomly
\State $f^{*}_{d_i} \gets \mathbf{TS}$(tgtDim=$d_i$, Fixed=$A[-d_i]$) and $A[d_i] = f^{*}_{d_i}$
\EndFor
\State Return $A = [f^{*}_{1:D}]$
\EndProcedure \label{destend}
\Procedure{\textbf{Boosted Destination Shift}}{} \label{boost}
\State Initial $A$ randomly
\For {$k = 1 \to K$}
\State Choose $d_i \in [1:D]$ randomly
\State $f^{*}_{d_i} \gets \mathbf{bstTS}$(tgtDim=$d_i$, Fixed=$A[-d_i]$) and $A[d_i] = f^{*}_{d_i}$
\EndFor
\State Return $A = [f^{*}_{1:D}]$
\EndProcedure \label{boostend}
\Function{TS}{tgtDim = $d_{i}$, Fixed=$[f_{d_{1}:d_{L}}]$ }
\For {$f_{d_{i}} = 1 \to N$}
\State $\theta^{f_{d_{i}}} \gets \mathbf{Sample}($Fixed=$[f_{d_{1}:d_{L}}, f_{(d_{i})}])$
\label{ts_sample}
\EndFor
\State \textbf{Return} $f^{*}_{d_{i}} \gets \argmax_{f_{d_{i}}}(\theta^{f_{d_{i}}})$
\EndFunction
\Function{bstTS}{tgtDim = $d_{i}$, Fixed=$[f_{d_{1}:d_{L}}]$ }
\For {$f_{d_{i}} = 1 \to N$}
\State $\mu_{d_{i}}^{1} \gets \mathbf{Sample}($Fixed=$[f_{d_{i}}])$
\For {$d_j = d_{1} \to d_{L} $}
\State $\mu_{d_i,d_{j}}^{2} \gets \mathbf{Sample}($Fixed$=[f_{d_{i}}, f_{d_{j}}])$
\EndFor
\State $\theta^{f_{d_{i}}} \gets \mu_{d_{i}}^{1} + \sum_{j=1}^{L} \mu_{d_i,d_{j}}^{2}$
\EndFor
\State \textbf{Return} $f^{*}_{d_{i}} \gets \argmax_{f_{d_{i}}}(\theta^{f_{d_{i}}})$
\EndFunction
\end{algorithmic}
\end{algorithm}

\textbf{Partial Path Finding:} 
In contrast, \textbf{PPF} describes a BFS alike algorithm. A $m$\ts{th}-partial path finding (\textbf{PPF$m$}) first recursively applies TS policy from nodes/dimensions under sequence $[d_{1}:d_{m-1}]$ (above level $m$ in graph view fig.~\ref{dg1}). Then it simultaneously visits the remaining $(D-m+1)$ dimensions (un-visited) in parallel at level $m$ and apply TS policy by keeping $[f^{*}_{d_{1}:d_{m-1}}]$ fixed correspondingly. Specifically, the \textbf{D-MABs} method is equivalent to \textbf{PPF1}, and it assumes independence among dimensions. The Pseudo code in algorithm~\ref{alg2} between line \ref{partial} and \ref{partialend} illustrates a \textbf{PPF2} algorithm, and it assumes conditionally independence among dimensions ($d_{j}$) given optimized content in dimension $d_{i}$. Mathematically, variables $A$ and $B$ are conditionally independent give $C$ ($A \perp B | C$) if and only if $P(A,B|C) = P(A|C) P(B|C)$, and would call joint distribution of ($A$, $C$) and ($B$, $C$) are independent. So the conditionally independence in \textbf{PPF2} also suggests independence among joint distribution of pairwise-dimensions. That is to say, \textbf{PPF2} models pairwise interactions between dimensions, as it draws samples from joint distribution of pairwise-dimensions. Generally \textbf{PPF$m$} models up to $(m)$-way interactions in tree model. The optimal computational complexity of \textbf{PPF2} is $O(SND)$ with $S$ times searching and space complexity is $O(ND^2)$, if only load hidden states from top 2 levels of the graph into memory.

\textbf{Destination Shift:} 
\textbf{DS} (algorithm~\ref{alg2} line \ref{dest}-\ref{destend}) is a naive adoption of hill-climbing method (red arrow in fig.~\ref{dg2}). It randomly inits an arm/layout $A$ (with equal chance) and performs hill-climbing method cycling through all dimensions for $K$ rounds. At each round $k$, it optimizes dimension $d_i$ chosen at random and returns the best  optimized content $f^{*}_{d_i}$ by TS with prefixed ($A[-d_i]$) in all other dimentions. Here $A[-d_i]$ means removal only the content in dimension (sub-index) $di$: $[f_{d_1:d_{i-1}}, f_{d_{i+1}:d_{D}}]$. The content of $A$ in $d_{i}$-th dimension is updated and assigned with $f^{*}_{d_i}$: $A[d_i] = f^{*}_{d_i}$. The computational complexity is $O(SNK)$ and space complexity is $O(N^D)$.

\textbf{Boosted Destination Shift:} 
Compare with \textbf{DS}, \textbf{Boosted-DS} employs \textbf{bstTS} instead of \textbf{TS} function for value optimization on each target node/dimension $d_i$. It extends our previous reasoning that sampling from joint distribution of $m$-dimensions is 1-to-1 mapping to regression model with m-way interaction. Pseudo code in Algorithm \ref{alg2} between line \ref{boost} and \ref{boostend} describes \nth{2}-Boosted-DS (\textbf{Boosted-DS2}) which models pairwise ($2-way$) interaction among dimensions. The reader could also view it in graph at fig.~\ref{dg2} in blue arrows. The difference btween textbf{TS} and \textbf{bstTS} function is that: textbf{TS} implicitly samples a single draw of each content $f_{d_{i}}$ within target dimension $d_i$ by fixing $[f_{d_{1}:d_{L}}]$ ($A[-d_i]$) at round $k$; but \textbf{bstTS} sums up multiple draws from $1$-way joint density ($P(f_{d_i})$) and all joint distributions of pairwise-dimensions ($2-way$) ($P(f_{d_i}, A[d_j] = f_{d_{j}})$. Generally, $m$\ts{th}-Boosted-DS (\textbf{Boosted-DSm}) would take the cummulative sums of drawing samples from joint distributions of upto $m-way$ dimensions . The computational complexity of \textbf{Boosted-DS2} is $O(SKND)$ and space complexity is $O(ND^2)$ if we store all needed hidden states instead. 

In summary, \textbf{PFP} and \textbf{Boosted-DS} are enhanced version of \textbf{FFP} and \textbf{DS}. \textbf{DS} and \textbf{Boosted-DS} randomly initialize a layout $A$ at beginning and keep improving itself dimension by dimension till converge, while \textbf{FFP} and \textbf{PFP} do not randomly initialize other dimension content. All four algorithms approximate the process of finding the best bandit arm by pruning search path and greedy optimization of the sequential process through all dimensions. As the greedy approach significantly reduce search space, the converge performances are expected to beat the traditional Thompson sampling method. 

\section{Empirical Validation and Analysis}\label{sec5}
We examine the performance of \textbf{FPF}, \textbf{PPF}, \textbf{DS} and \textbf{Boosted-DS} on simulated data set, comparing with \textbf{MVT}\cite{amazon2017}, \textbf{$\mathbf{N^D-}$ MAB}\cite{amazon2017} and \textbf{D-MABs}\cite{amazon2017} other models mentioned before. Specifically, we evaluate (a) the average cumulative regret, (b) the convergence speed, and (c) the efficiency of best optimal arm selection among these models under the same simulation environment settings. To access fairly appreciative analysis, the simulator's mechanism and parameters are randomly initialized and reset for each new experiment setting. We also replicate all simulation H(=100) times and take the average to eliminate evaluation bias due to sampling randomness. Furthermore, we extensively check the cumulative regret performance of proposed algorithms by varying (1) the relative strength of the interaction between dimensions and (2) complexity in arm space (altering $N$ and $D$) to gain a comprehensive understanding of our model. In the end, we conduct an off-policy evaluation on real data sets.

\begin{table}[htbp]
\centering
\begin{tabular}{@{}llll@{}}
\toprule
Algorithm & Description & Parameters & Speed \\ \midrule
\textbf{$N^D$-MAB} & A bandit ($N^D$-arms) & $O(N^D)$ & 28.14 it/s \\
\textbf{D-MAB} & D bandits (N-arms) & $O(ND)$ & 63.08 it/s \\
\textbf{MTV2} & Probit linear & $O(N^2D^2)$ & 0.25 it/s \\
\textbf{FPF} & Full Path Finding & $O(N^D)$ & 7.04 it/s \\
\textbf{PPF2} & Partial Path Finding & $O(ND^2)$ & 22.39 it/s \\
\textbf{DS} & Destination Shift & $O(N^D)$ & 2.01 it/s \\
\textbf{Boosted-DS2} & Boosted Dest. Shift & $O(ND^2)$ & 1.38 it/s \\ \bottomrule
\end{tabular}
\caption{Iteration speed of algorithms discussed} \label{iteration}
\end{table}

\subsection{Simulation Settings}
The Bernoulli reward simulator is based on linear model with $m$-way interactions:
\begin{equation}\label{simul}
\scalemath{0.7}{
\Phi(\theta) = \frac{1}{\beta} [ \mu^{0} + \alpha_1 \sum_{i = 1}^{D} \mu_{i}^{1}(A) + \dots + \alpha_m \sum_{d_1 = 1}^{D}\dots\sum_{d_m = d_{m - 1} + 1}^{D} \mu_{d_1,\dots,d_m }^{m}(A) ]
}
\end{equation}
where $\beta$ is scaling variable and $\alpha_1$, $\dots$, $\alpha_m$ are control parameters, $\mu_{d_1,\dots,d_m }^{m}(A)$ represents the weight of $m$-way interaction among $d_1,\dots,d_m$. It is trivial to set $\mu^{0} =0$. The weights \textbf{$\mathbf{\mu}$} are initialized from $N(0, 1)$ randomly and independently. We further set $\beta = m$ and $\alpha_m = \frac{m!}{D(D-1)\dots(D-m+1)}$ to control the overall signal to noise ratio as well as related strength among $m$-way interactions.

In this paper, we set $m=2$ (pairwise dimension interaction), $D = 3$ and $N = 10$ in above simulator settings, which yields $1000$ possible layouts. To observe the convergence of each model and eliminate the randomness, our simulation is iterated with $T = 100, 000$ steps per experiment and duplicated $H = 100$ replications. On each experiment replica and at time step $t$, layout $A^{t}$ is chosen by each algorithm, and a binary reward $X^{t}$ is sampled from Bernoulli simulator with success rate $\theta$, which is formulated from Equation \ref{simul} with random initial weights \textbf{$\mu$}. We choose $S = 45$ and $K = 10$ as the same hill-climbing model parameter settings in different algorithms. 

\subsection{Numerical Results} 
Fig.~\ref{hist} shows histograms of arm exploration and selection activities for $7$ algorithms plus distribution of simulator success rates. The horizontal axis is the success rate of the selected arm, while the vertical axis is the probability density. The Bernoulli simulators are symmetrically distributed, which coincides with random initial settings. Meanwhile, the severity of right skewness reveals the efficiency of recognizing bad performance arms and quickly adjusting search space to better arms. Though \textbf{$\mathbf{N^D}$-MAB} is theoretically guaranteed to archive optimal performance in long run, the histograms empirically suggest TS-PP (\textbf{MVT2}, \textbf{FPF}, \textbf{PPF2} and \textbf{Boosted-DS2}) outperform \textbf{$\mathbf{N^D}$-MAB} in many ways. The efficiency performance of \textbf{DS} is worse than \textbf{Boosted-DS2} and \textbf{PPF2}. The underlying reason can be that only a small fraction of arms are explored at an early stage, and little information is known for each arm. Starting from top strategy can resemble dimensional analogs and learn arm's reward distribution from other arms with similar characteristics. In turn, it shifts a heuristic search to better-performed arms rapidly. \textbf{Boosted-DS2} combats \textbf{DS}'s weakness using TS samples from top levels to mimic samples drawn from the lower. 

\begin{figure}[htbp]
\centering
\includegraphics[width=\linewidth, scale = 0.8]{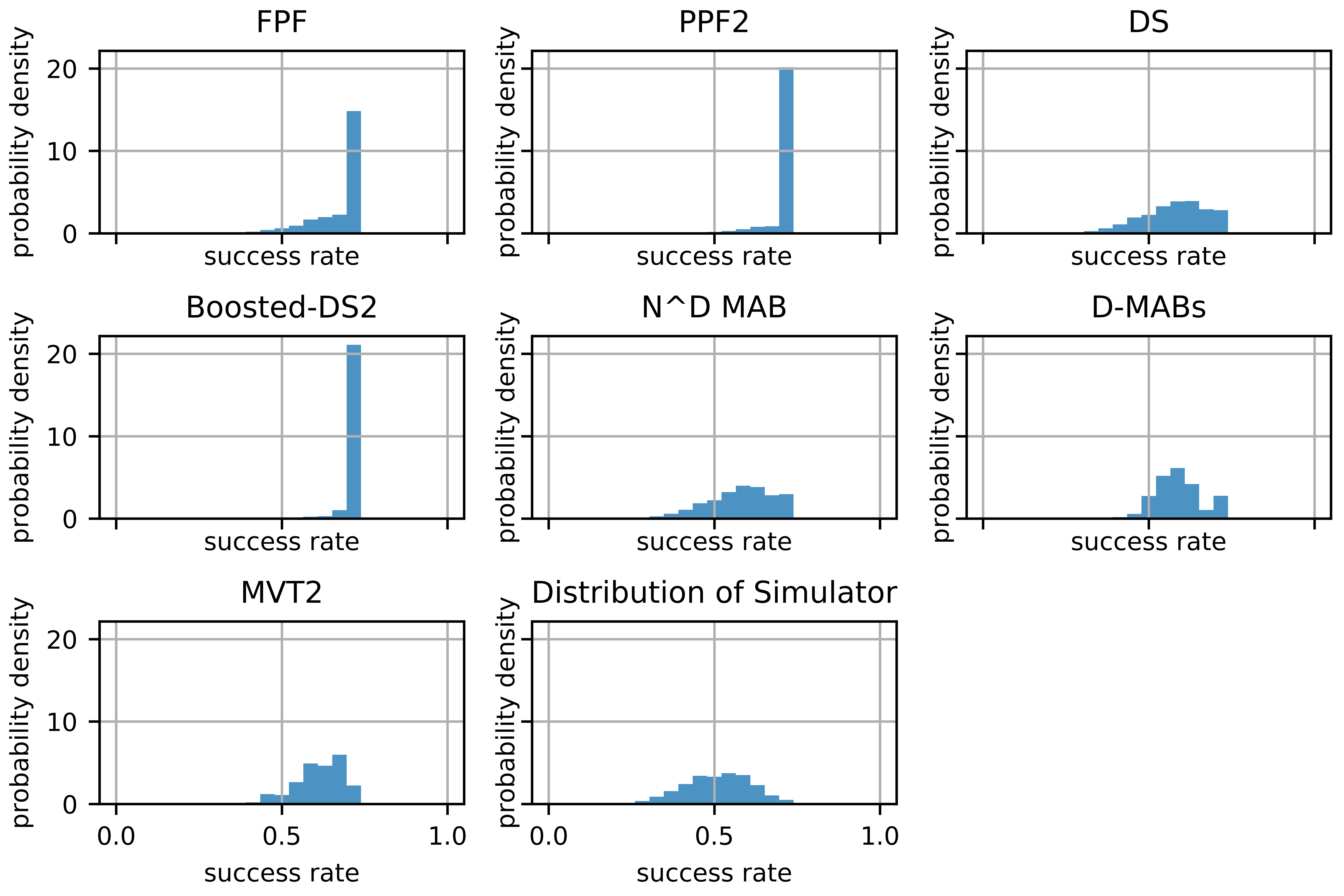}
\caption{Histogram of expected reward for historical arm search.}
\label{hist}
\end{figure}

We use Average Regret, Convergence Rate, and Best Arm Rate as our evaluation criteria. We define Convergence Rate as the proportion of trials with the most selected layout over a moving window with batch size $t_1 - t_0=$ 1000. We further specify Best Arm Rate as the proportion of trials with the true best layout in a batch. 

\begin{equation*} 
\scalemath{0.9}{
\begin{aligned}
& \mathbf{Average \: Regret} = \sum^{H}_{h = 1}\frac{1}{H}\sum^{T}_{t = 1} \frac{E(X_{A^{\star}}) - X^{t} }{T}, \\
& \mathbf{Convergence \: Rate} [t_0, t_1] = \sum^{H}_{h = 1}\frac{1}{H}\sum^{t_1}_{t = t_0} \frac{\mathbf{1}_{A^{\text{o}}}(A^{t})}{t_1 - t_0}, \\
& \mathbf{Best \: Arm \: Rate} [t_0, t_1] = \sum^{H}_{h = 1}\frac{1}{H}\sum^{t_1}_{t = t_0} \frac{\mathbf{1}_{A^{\star}}(A^{t})}{t_1 - t_0},
\end{aligned}
}
\end{equation*}
where $A^{\text{o}}$ and $A^{\star}$ stand for most often selected arm and best possible arm respectively within a batch. The ideal convergence rate and best arm rate shall both reach to 1, which means algorithm converges selection to the best arm. In practice, fully converged trials almost surely select the same (sub-optimal) but not necessarily the best. 

Simulation results are displayed in fig.~\ref{performance} where the x-axis is the time steps. Path planning algorithms demonstrate advantages over base models, especially for \textbf{FPF}, \textbf{PPF2} and \textbf{Boosted-DS2}. We see that \textbf{PPF2} and \textbf{Boosted-DS2} quickly jump to low regret (and high reward) within $5000$ steps, followed by \textbf{FPF} and \textbf{MVT2} around $10000$ steps. Although \textbf{Boosted-DS2} and \textbf{MVT2} share fastest convergence speed followed by \textbf{PPF2} then \textbf{FPF}, but \textbf{FPF} holds the highest best arm rate. \textbf{FPF} performance in cumulative regret (and reward) catches up for longer iterations as well. The rationale behind these is that \textbf{FPF} involves the most complicated model space with considering full dimension interactions, in which it not only looks from the higher level of the graph to quickly eliminate badly performed dimension contents but also drills down to leaf arms to correct negligence from higher levels. The exponential space complexity or computational complexity proportional to $T$ is our concern of \textbf{FPF} compared with \textbf{PPF2} and \textbf{Boosted-DS2}.

\begin{figure}[htbp]
\centering 
\begin{subfigure}[b]{0.9\columnwidth} 
\includegraphics[width=\columnwidth]{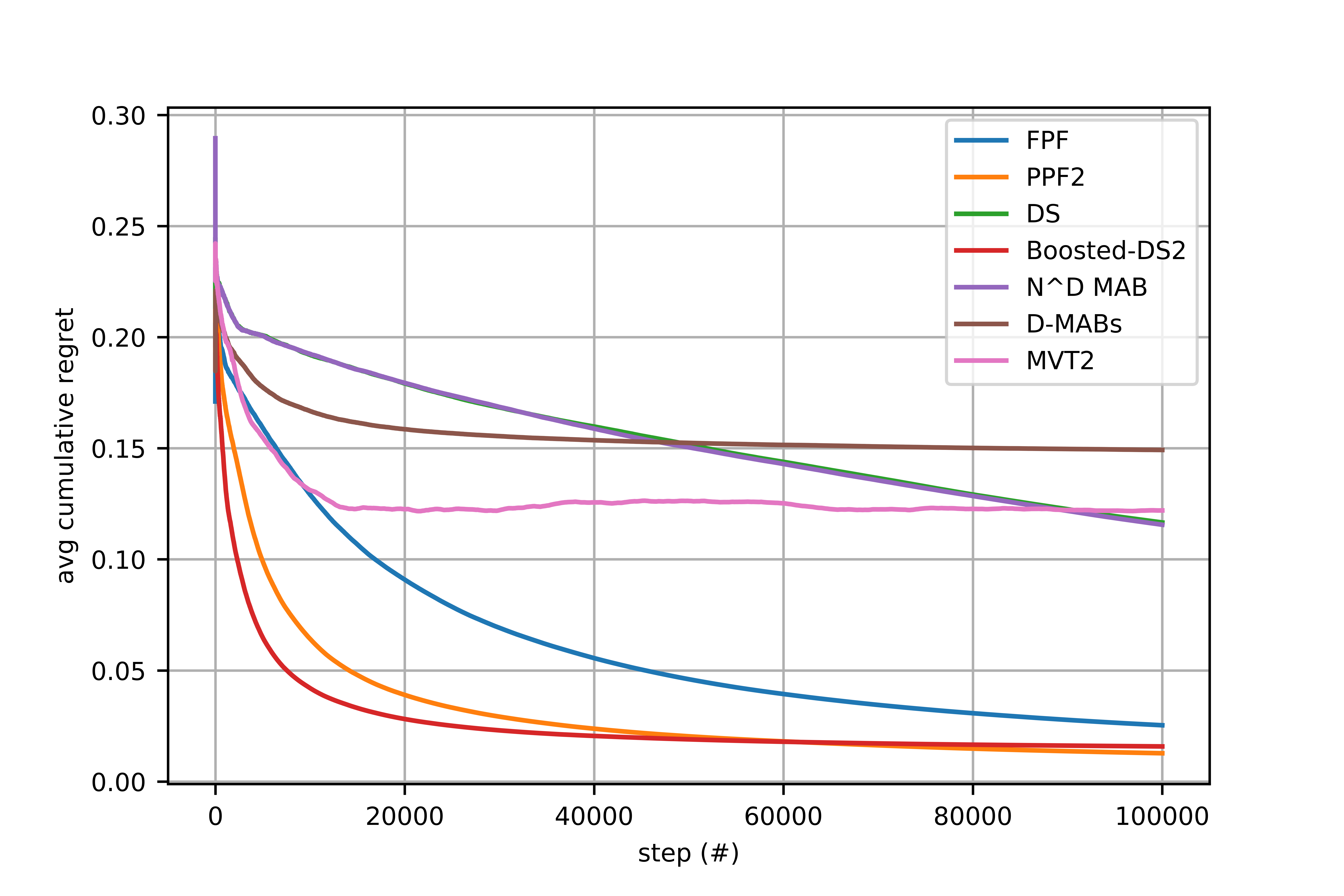}
\caption{Average Regret.}
\label{pf1}
\end{subfigure}  
\begin{subfigure}[b]{0.9\columnwidth}
\includegraphics[width=\columnwidth]{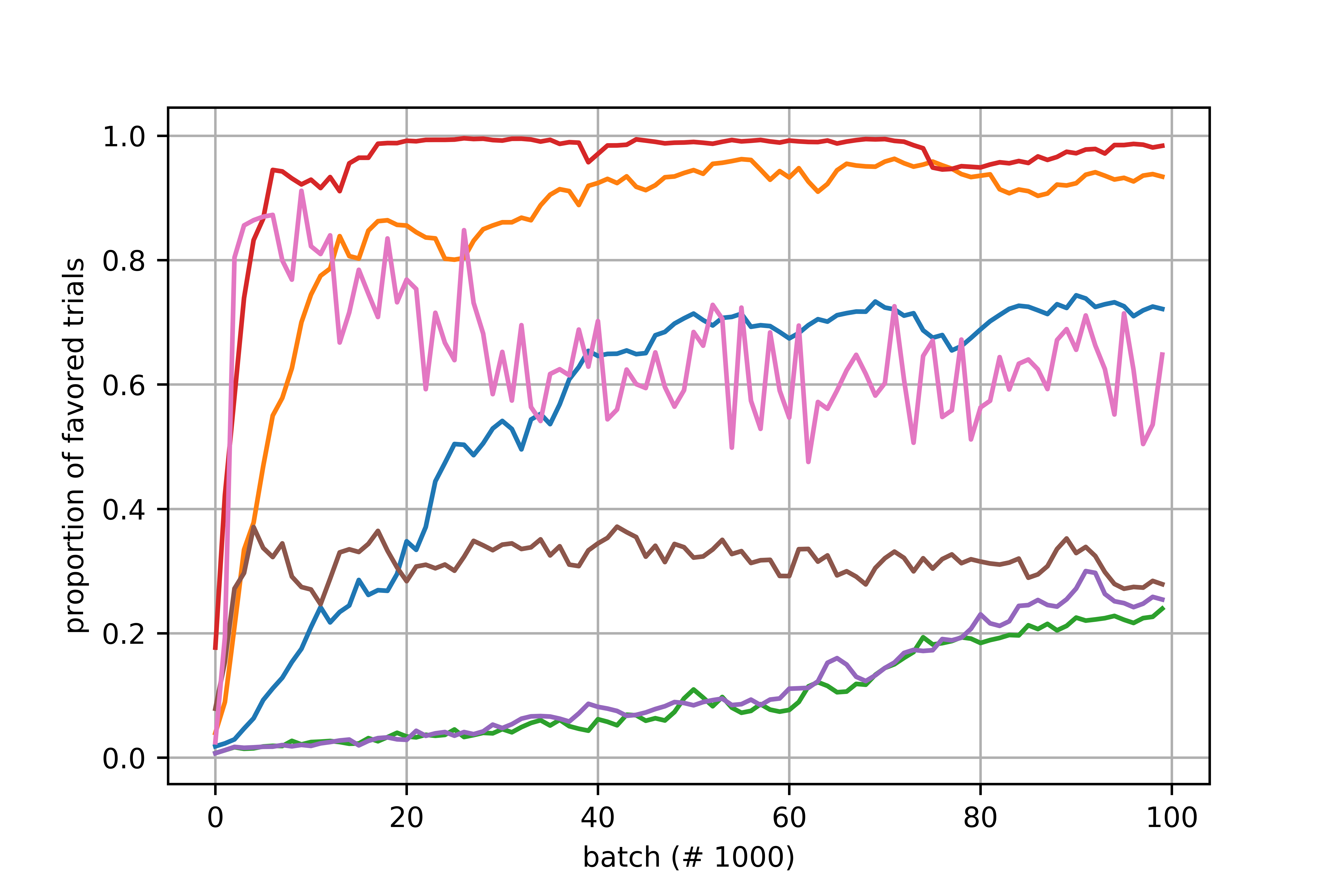}
\caption{Convergence Rate}
\label{pf3}
\end{subfigure} 
\begin{subfigure}[b]{0.9\columnwidth}
\includegraphics[width=\columnwidth]{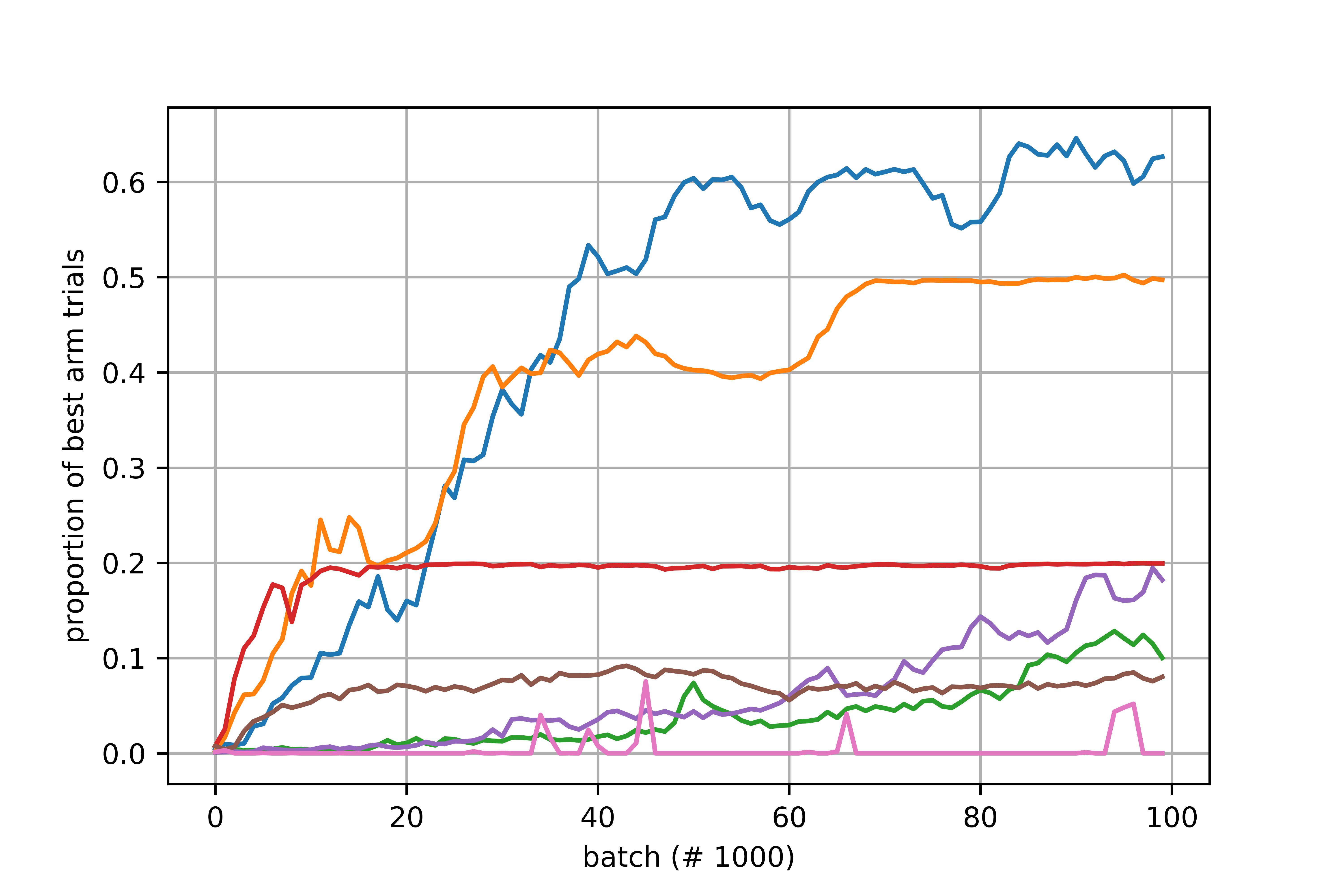}
\caption{Best Arm Rate}
\label{pf4}
\end{subfigure}
\caption{Performance of algorithms on simulated data with $D = 3, N = 10, \beta = 3$, $m = 2$, $\alpha_1 = \frac{1}{3}$ and $\alpha_2 = \frac{1}{3}$. }
\label{performance}
\end{figure}

In our experiment, \textbf{PPF2}, \textbf{Boosted-DS2}, and \textbf{MVT2} assume models with pairwise interactions, and it happens to be our simulator setting. In practice, extra effort is needed for correct modeling of the reward function, which is out of this paper's scope. \textbf{PPF2} and \textbf{Boosted-DS2} both efficiently receive lower regret and better best arm rate comparing with \textbf{MVT2}. However, \textbf{PPF2} carries better best arm rate than \textbf{Boosted-DS2}. Our takeaway here is that the hill-climbing strategy starts with randomly guess which easily ends up with local optimal (low regret and high convergence), not the global best optimal (low best arm rate). Meanwhile, \textbf{D-MABs} struggles in performance as its independence among dimensions assumption contradict with our simulator. 

Though \textbf{PPF2}, \textbf{Boosted-DS2} and \textbf{MVT2} have similar complexity in parameter space, \textbf{MVT2} demands longer time per iteration than others. Table \ref{iteration} shows speed of our implementation. Both in theoretical and practical, \textbf{MVT2} is the slowest one as the heavy computation burden when updating the posterior sampling distribution of regression coefficients. 

\subsection{Change Experiment Settings}

\begin{figure}[ht]
\centering 
\begin{subfigure}[b]{0.9\columnwidth} 
\includegraphics[width=\columnwidth]{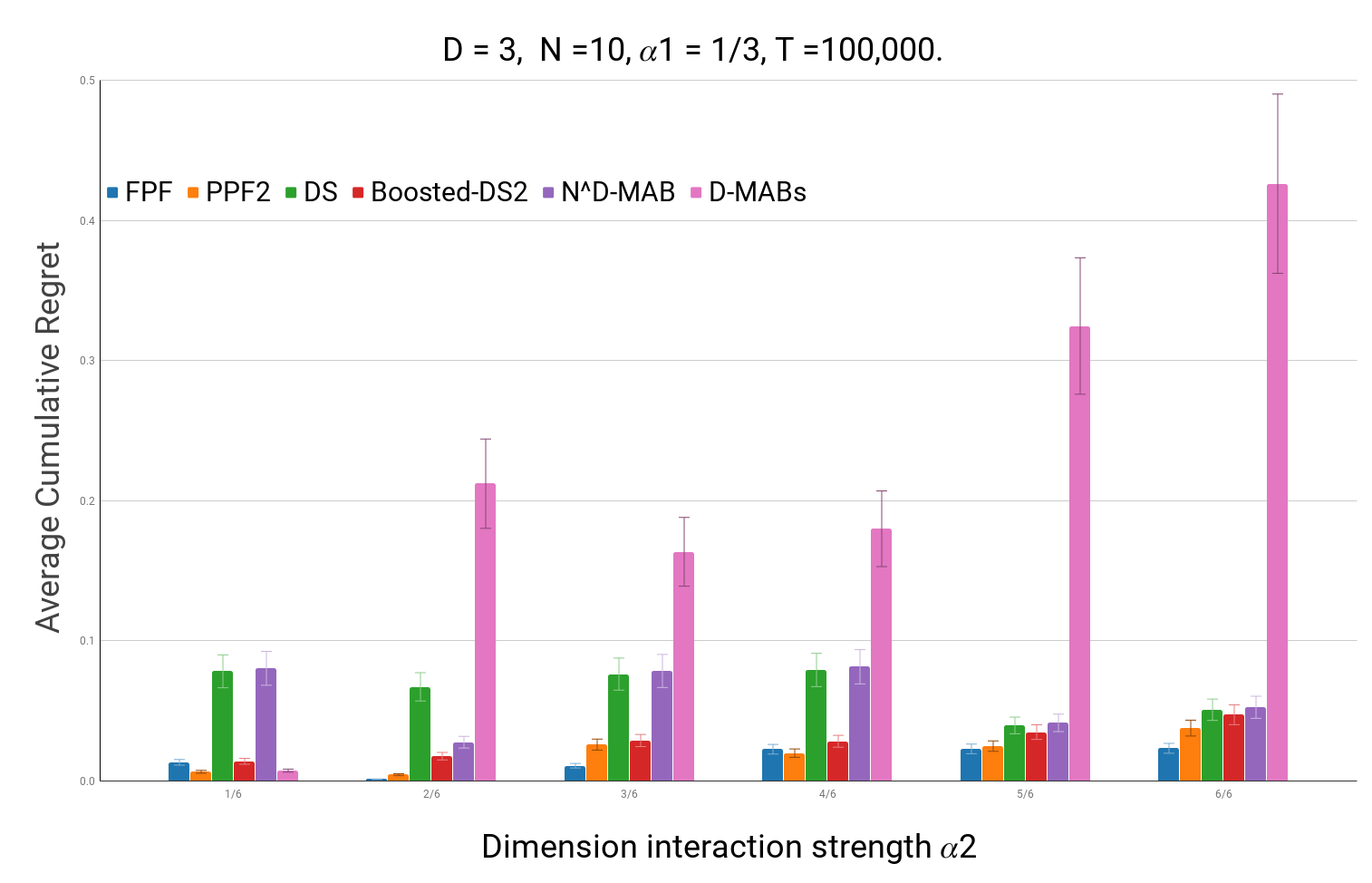}
\caption{Algorithm performance when $\alpha_2$ varies}
\label{varpf1}
\end{subfigure} 
\begin{subfigure}[b]{0.9\columnwidth} 
\includegraphics[width=\columnwidth]{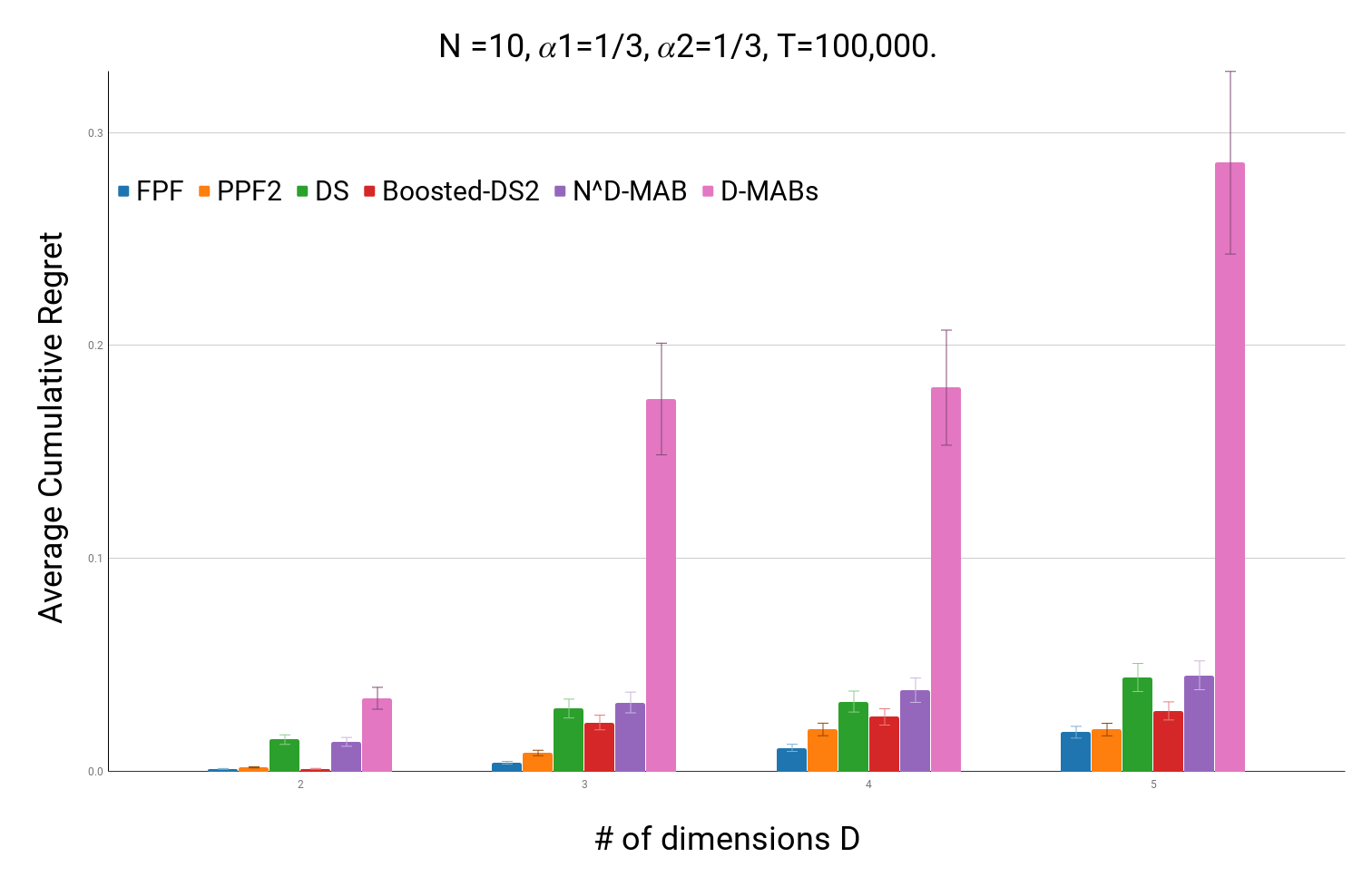}
\caption{Algorithm performance when $D$ varies}
\label{varpf2}
\end{subfigure} 
\begin{subfigure}[b]{0.9\columnwidth} 
\includegraphics[width=\columnwidth]{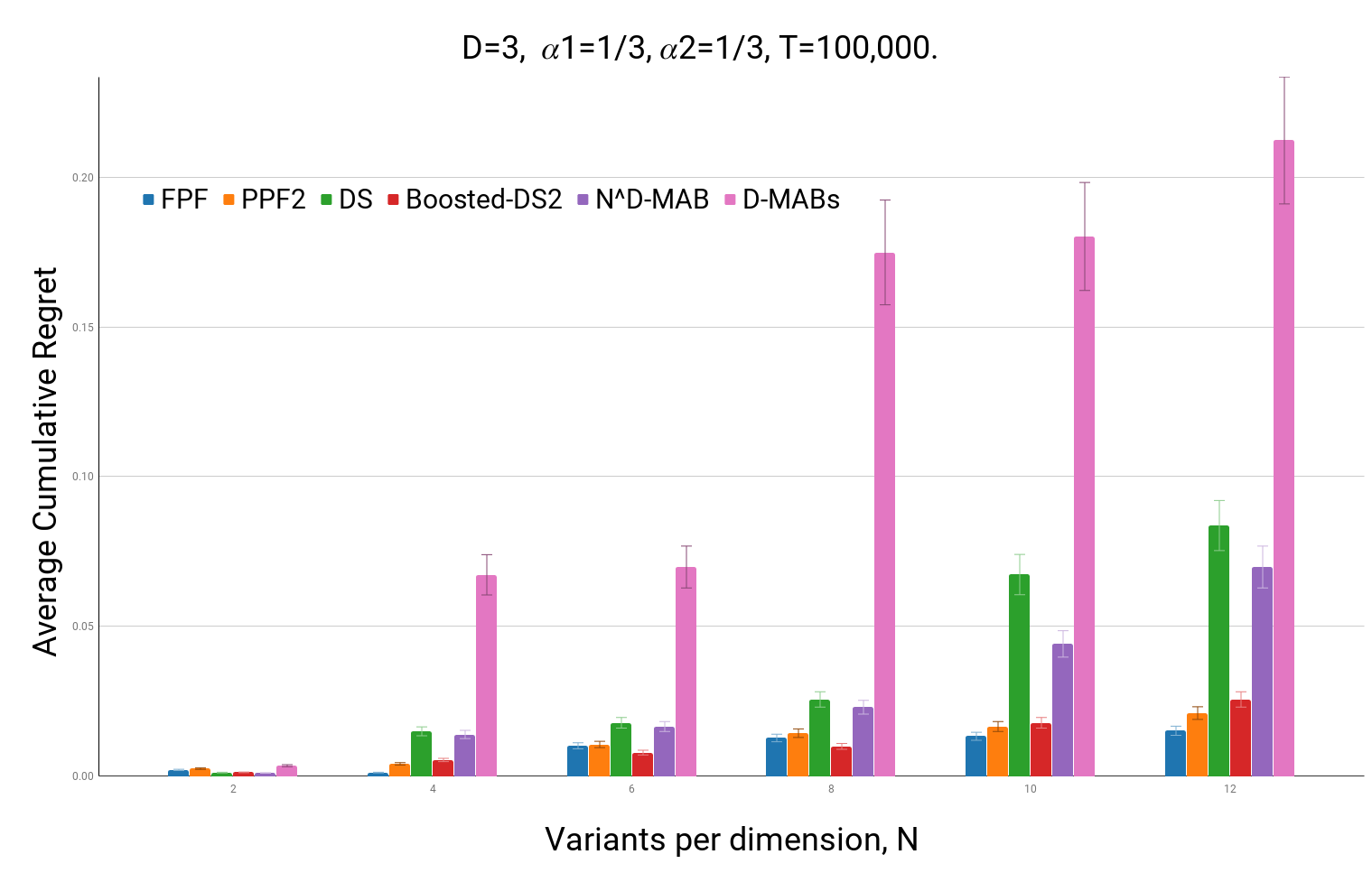}
\caption{Algorithm performance when $N$ varies}
\label{varpf3}
\end{subfigure} 
\caption{Performance of algorithms with standard error when D, N and $\alpha_2$ varies. Average regret and standard error are computed based on H = 20 iterations. } 
\label{varperformance}
\end{figure}

We re-run our experiments and check average cumulative regret performance with varying $\alpha_2$ to change the strength of interactions as well as varying $N$ and $D$ to change space complexity in fig.~\ref{varperformance}. We skip $\textbf{MVT2}$ due to time-consuming ($\textbf{MVT2}$ takes 5 days per experiment per iteration). Each new experiment iterates T = 100000 steps and repeats H = 20 times. Standard errors are also provided in fig.~ \ref{varperformance}. As $\alpha_2$ varies from $\frac{1}{6}$ to 1 with step size $\frac{1}{6}$, we see the pattern consists with prior result at fig.~\ref{varpf1}.
 The only exception is $\textbf{D-MABs}$ gets impressing regret when interaction strength is weak enough ($\alpha_2 = \frac{1}{6}$), due to $\textbf{D-MABs}$'s no interaction assumption. Next we analyze how the searching space ($N^D$) could impact performance. We systematically swap $N$ in $(2, 4, 6, 8, 10, 12)$ and $D$ in $(2, 3, 4, 5)$ at fig.~\ref{varpf2} and \ref{varpf3} respectively. It shows that the previous result still holds: $\textbf{FPF} \; \simeq \; \textbf{PPF2} \; \simeq \; \textbf{Boosted-DS2} \; > \; \textbf{DS} \; \simeq \; \textbf{$\mathbf{N^D}$-MAB} \; > \; \textbf{D-MABs}$. Based on extensive experiments, we assert that our proposed method is superior consistently.

\subsection{Off-Policy Evaluation on Real Data}

The two validation datasets described in Table \ref{tb2} are coming from eBay A/B testing experiments for layout optimization and public available on the LIBSVM website \footnote{\url{https://www.csie.ntu.edu.tw/~cjlin/libsvmtools/datasets/binary.html}}. We choose estimator method mentioned in \cite{langford2008exploration} as our off-policy evaluation method. Instead of naive estimation, we adopt James-Stein estimator \cite{dimmery2019shrinkage, efron1971limiting, efron1975data} being our estimation for each arm. Off-policy evaluation is conducted after 100000 iterations/steps with 100 repetitions and plotted with standard errors.

\begin{table}[htbp] 
\centering
\begin{tabular}{@{}llll@{}}
\toprule
Datasets & Source & $\#$ of samples & $\#$ of features \\ \midrule
four-class (binary) & LIBSVM & 863 & $4 \times 4$ \\ 
A/B testing data & eBay & 88M & $2 \times 3 \times 4$ \\ \bottomrule
\end{tabular}%
\caption{Datasets characteristics} \label{tb2}
\end{table}

\begin{figure}[htbp]
\centering 
\begin{subfigure}[b]{0.9\columnwidth} 
\includegraphics[width=\columnwidth]{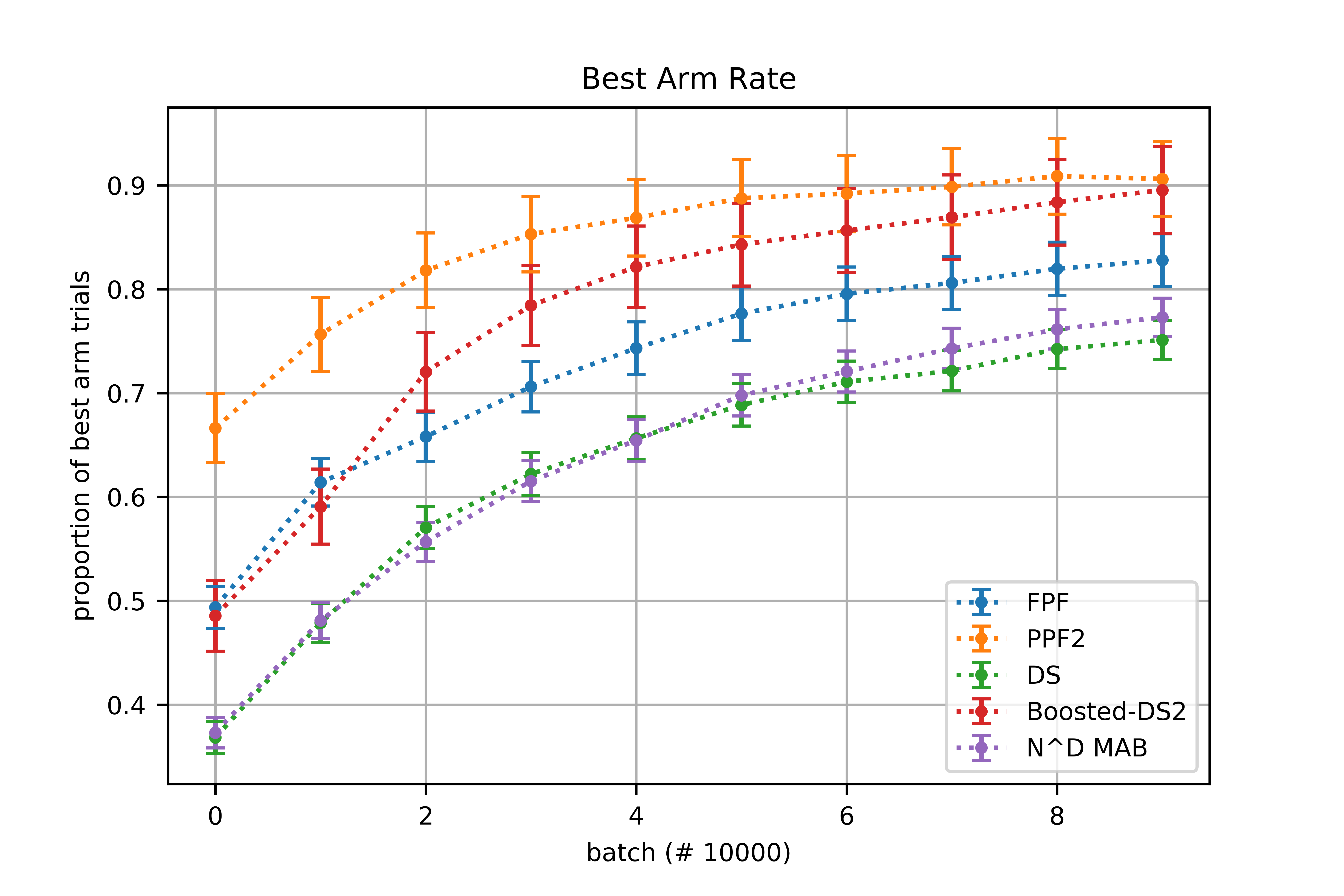}
\caption{Best arm rate for every 10000 steps}
\label{ebay1}
\end{subfigure} 
\begin{subfigure}[b]{0.9\columnwidth} 
\includegraphics[width=\columnwidth]{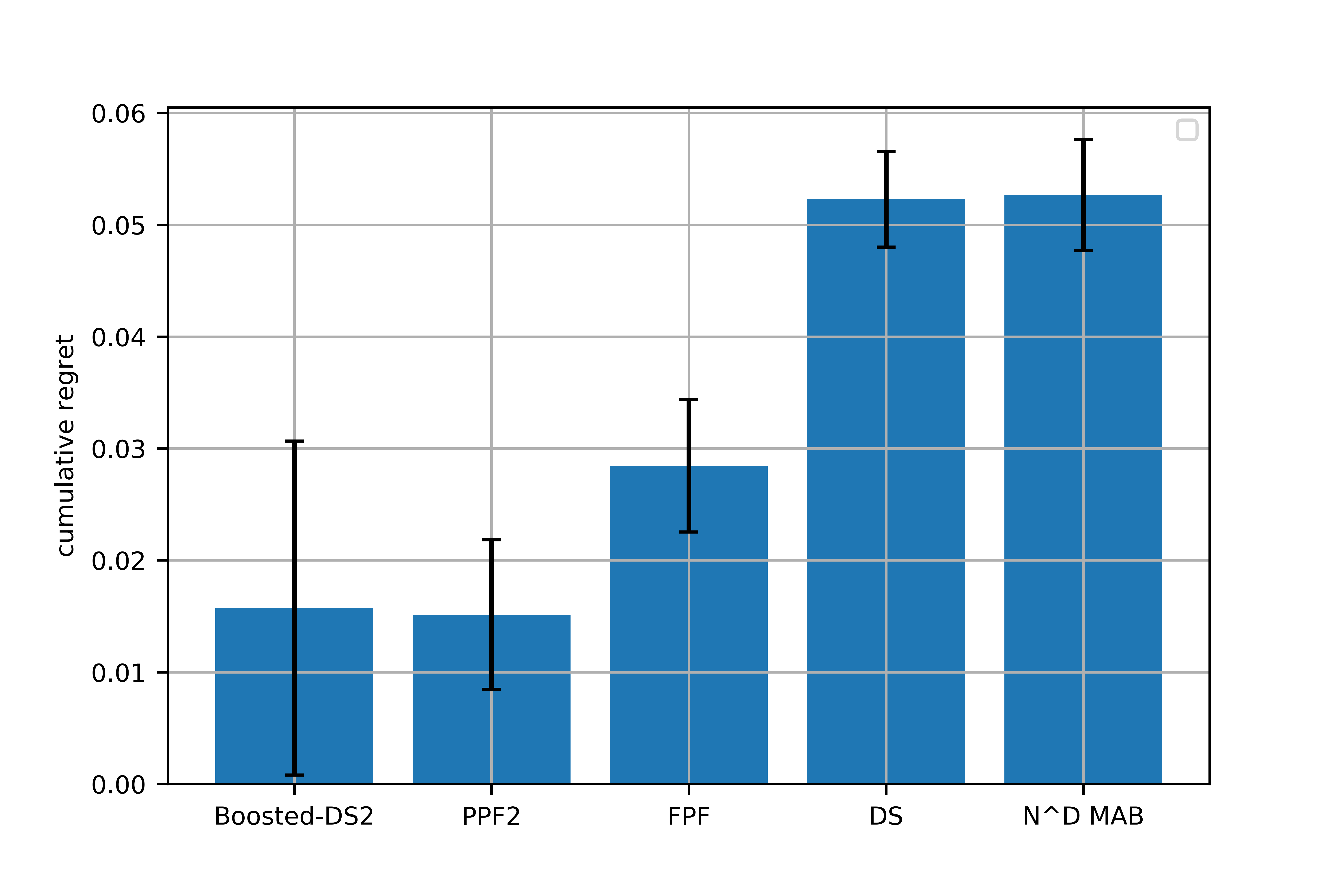}
\caption{Accumulated regret at 100000 step}
\label{ebay2}
\end{subfigure} 

\begin{subfigure}[b]{0.9\columnwidth} 
\includegraphics[width=\columnwidth]{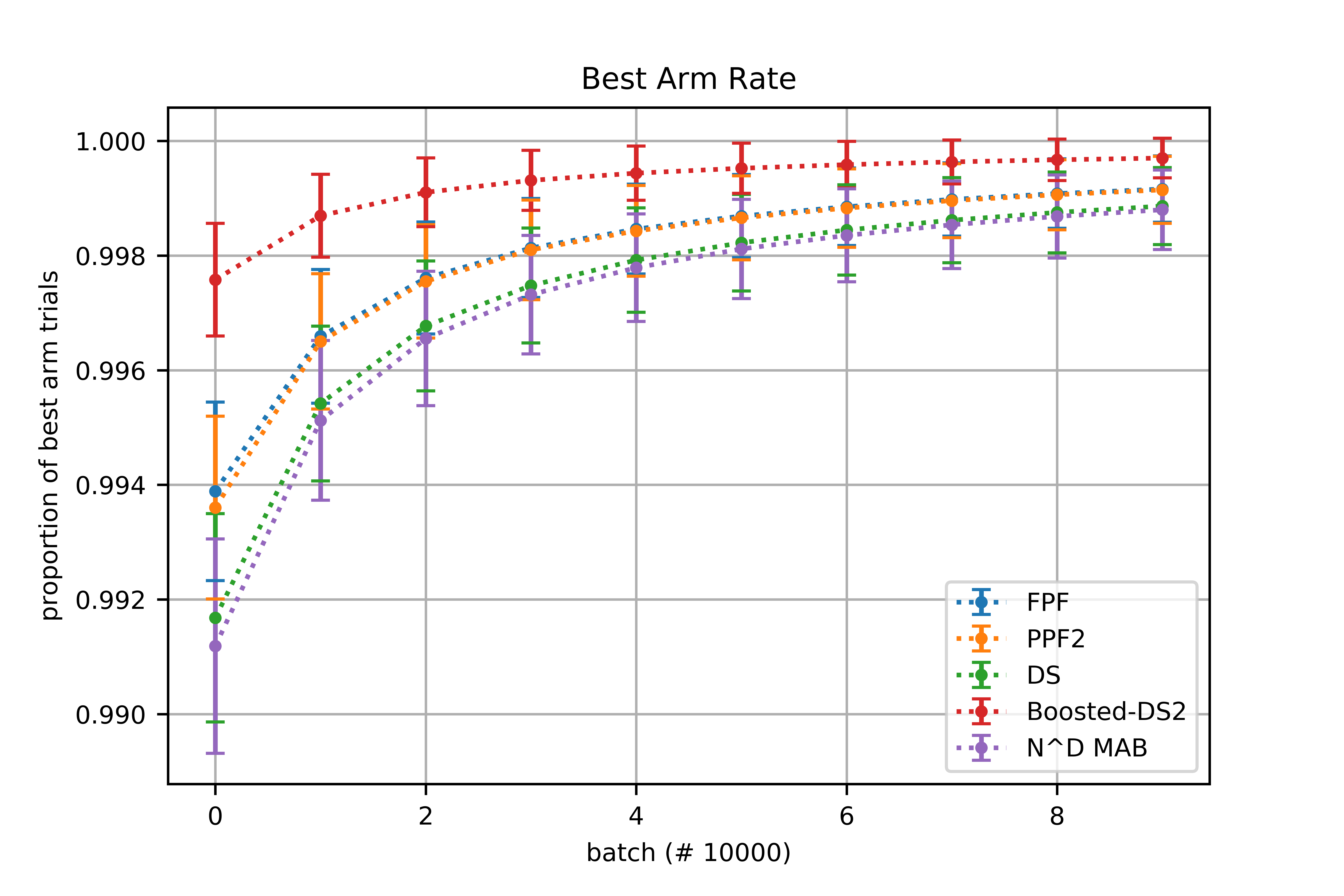}
\caption{Best arm rate for every 10000 steps}
\label{fourclass1}
\end{subfigure} 
\begin{subfigure}[b]{0.9\columnwidth} 
\includegraphics[width=\columnwidth]{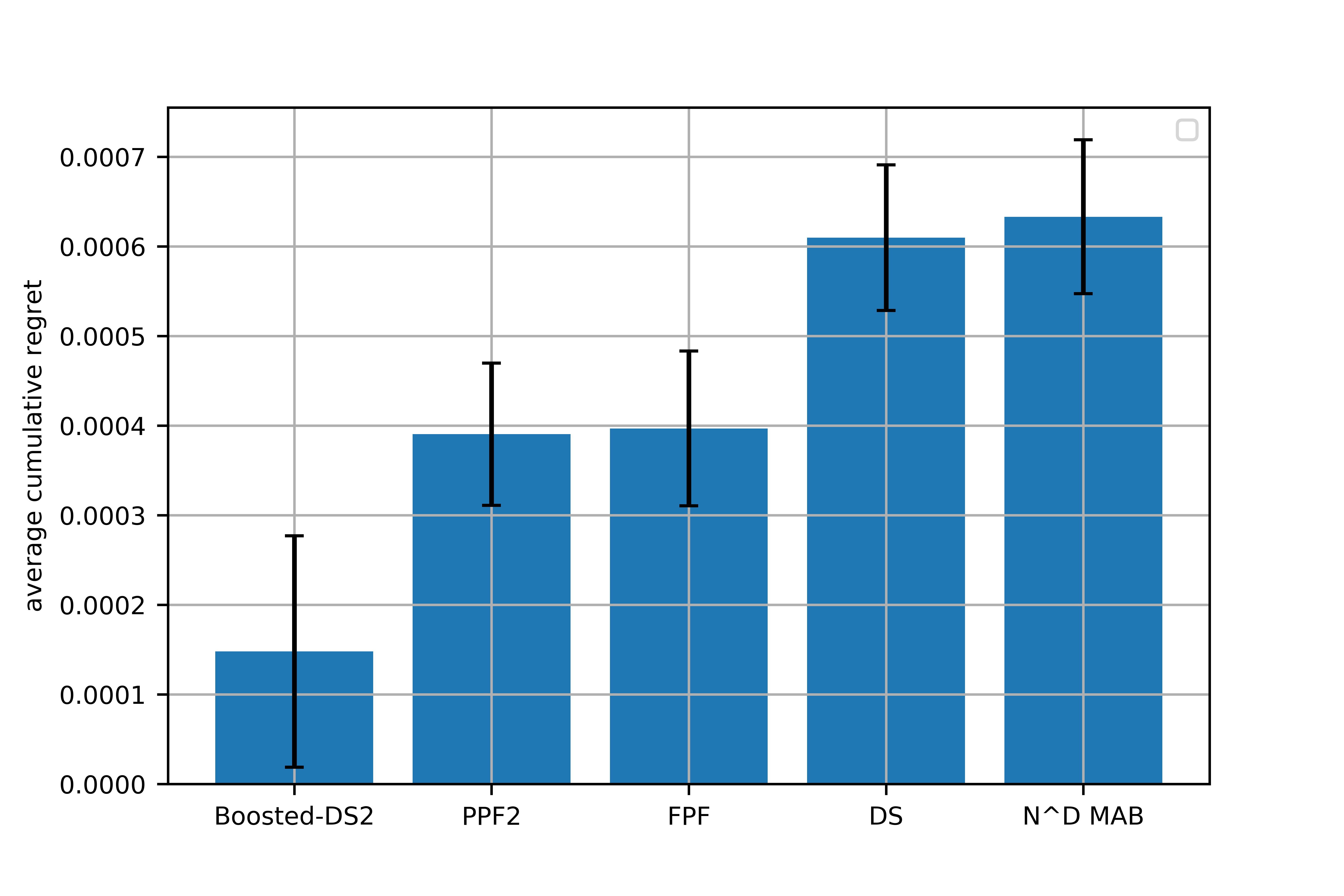}
\caption{Accumulated regret at 100000 step}
\label{fourclass2}
\end{subfigure} 
\caption{Off-policy evaluation on eBay past experiment data (\ref{ebay1}, \ref{ebay2}) and fourclass data (\ref{fourclass1}, \ref{fourclass2}). Best arm rate and regret are calculated at 10000 steps with 100 repetitions ans plot with standard errors. } 
\label{realdata}
\end{figure}

Our validation results on real data are plotted in fig.~\ref{realdata}. $\textbf{D-MABs}$ performs the worst, so we remove it to have better visualization in our comparison. We see that $\textbf{PPF2}$ and $\textbf{Boosted-DS2}$ on improved of $40\%$ to $70\%$ on average regret, comparing with \textbf{$N^D$-MAB}. The best arm rate graphs also confirm this claim. It is worth to mention that $\textbf{Boosted-DS2}$ has better best-arm rate and regret than $\textbf{PPF2}$ in four-class dataset. We suspect that it is due to the number of arms (16) is not huge.

Overall, our results suggest that TS-PP has good performance overall for multivariate bandit problems with large search space when the dimension hierarchy structure exists. \textbf{PPF} and \textbf{Boosted-DS} attract our attention due to its computation efficiency with remarkable performance.

\section{Conclusions}\label{sec6}
This paper presented TS-PP algorithms capitalizing on the hierarchy dimension structure in Multivariate-MAB to find the best arm efficiently. It utilizes paths of the hierarchy to model the arm reward success rate with m-way dimension interaction and adopts TS for a heuristic search of arm selection. It is intuitive to combat the curse of dimensionality using serial processes that operate sequentially by focusing on one dimension per each process. In both simulation and real case evaluation, it shows superior cumulative regret and converges speed on large decision space. The \textbf{PPF} and \textbf{Boosted-DS} stand out due to its simplicity and high efficiency.

It is trivial to extend our algorithm to contextual bandit problem with finite categorical context features. But how to extend our algorithm from discrete to continuous contextual variables worth further exploration. We notice some related work of TS-MCTS \cite{bai2014thompson} dealing with continuous rewards in this area. Finally, fully understanding the mechanism of using a greedy heuristic approach (in our method) to approximate TS from $N^D$ arms is still under investigation. 


\end{document}